\documentclass[final]{cvpr}
\usepackage{float}
\usepackage{times}
\usepackage{epsfig}
\usepackage{graphicx}
\usepackage{amsmath}
\usepackage{amssymb}
\usepackage{subfigure}
\usepackage{booktabs}

\usepackage[pagebackref=true,breaklinks=true,colorlinks,bookmarks=false]{hyperref}

\def\cvprPaperID{7839} 
\def\confYear{CVPR 2021}
\usepackage{microtype}

\usepackage{color}

\graphicspath{{figures/}}
\definecolor{lightgray}{gray}{.92}
\definecolor{tinygray}{gray}{.96}

\begin{document}

\title{Drafting and Revision: Laplacian Pyramid Network for Fast High-Quality \\ Artistic Style Transfer}

\author{Tianwei Lin$^1$, Zhuoqi Ma$^{1,2}$, Fu Li$^1$, Dongliang He$^1$, Xin Li$^1$, Errui Ding$^1$, \\
Nannan Wang$^2$, Jie Li$^2$, Xinbo Gao$^3$
 \\
Department of Computer Vision Technology (VIS), Baidu Inc.$^1$ \\
Xidian University.$^2$ 
Chongqing University of Posts and Telecommunications.$^3$\\
	{\tt\small lintianwei01@baidu.com, zhuoqi\_ma@hotmail.com } 
}


\twocolumn[{%
\maketitle
\begin{figure}[H]
\hsize=\textwidth 
\centering
\includegraphics[width=16.5cm]{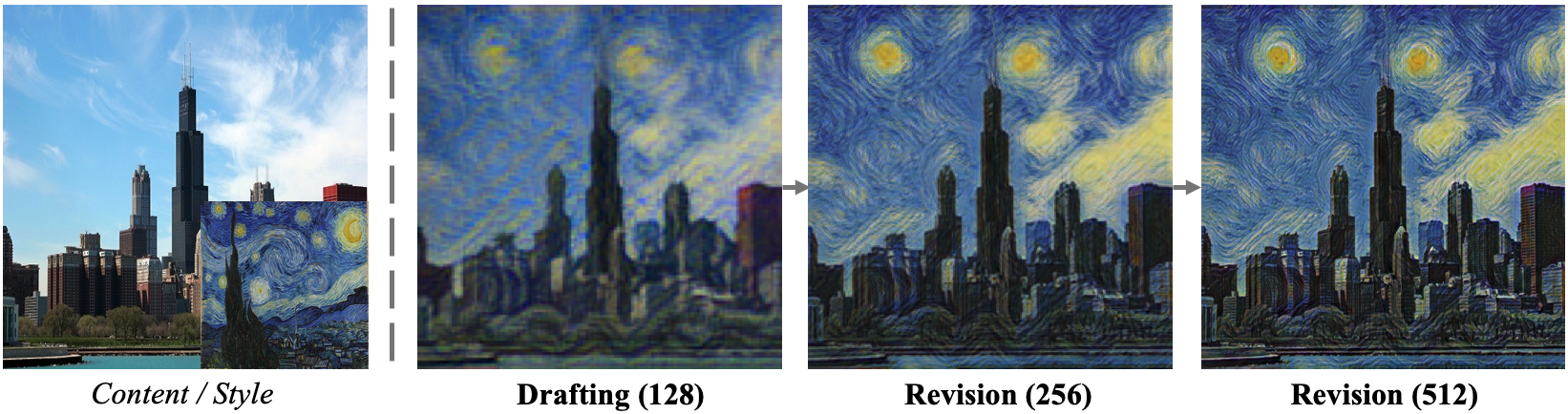}
\caption{Illustration of our proposed style transfer process. First we transfer global patterns in low resolution, then revise local patterns in high resolution. For better visualization, we resize stylized images of different scale into {\color{black}{the}} same size. Zoom in to have {\color{black}{a}} better view.}
\label{fig:overview}
\end{figure}
}]


  
	
\begin{abstract}
Artistic style transfer aims at migrating the style from an example image to a content image.
Currently, optimization-based methods have achieved great stylization quality, but expensive time cost restricts their practical applications. 
Meanwhile, feed-forward methods still fail to synthesize complex style, especially when holistic global and local patterns exist.
Inspired by the common painting process of drawing a draft and revising the details, we introduce a novel feed-forward method named \textbf{Laplacian Pyramid Network (LapStyle)}. {\color{black}{LapStyle first transfers global style patterns in low-resolution via a Drafting Network. It then revises the local details in high-resolution via a Revision Network, which hallucinates a residual image according to the draft and the image textures extracted by Laplacian filtering. Higher resolution details can be easily generated by stacking Revision Networks with multiple Laplacian pyramid levels. The final stylized image is obtained by aggregating outputs of all pyramid levels}}. 
Experiments demonstrate that our method can synthesize high quality stylized images in real time, where holistic style patterns are properly transferred.
Codes will be released on \href{https://github.com/PaddlePaddle/PaddleGAN/}{PaddleGAN}.
%

\end{abstract}
	
\section{Introduction}

Artistic style transfer is an attractive technique which can create an art image with the structure of a content image and the style patterns of an example style image.
{\color{black}{It has been a prevalent research topic for both academy and industry.}}
%
Recently, there have been a lot of methods proposed for neural style transfer, which can be roughly divided into two types: image-optimization and model-optimization methods.

Image-optimization methods {\color{black}{iteratively}} optimize {\color{black}{stylized}} image with fixed network. The seminal work of Gatys \emph{et al.} \cite{gatys2016image} achieves style transfer in an iterative optimization process, where the style patterns are captured by correlation of features extracted from a pre-trained deep neural network.
Following works improve \cite{gatys2016image} mainly in the form of different loss functions \cite{kolkin2019style, risser2017stable}.
Although superior stylization results are achieved, \eg, STROTSS \cite{kolkin2019style}, {\color{black}{widespread}} applications of these methods are still restricted by their slow online optimization process.
%
On the contrary, model-optimization methods update neural networks by training and are feed-forward in testing. There are mainly three subdivided types: (1) \emph{Per-Style-Per-Model} methods \cite{johnson2016perceptual, li2016precomputed, ulyanov2016texture, ulyanov2016instance,ulyanov2017improved} are trained to synthesize images with a single given style image; (2) \emph{Multi-Style-Per-Model} methods \cite{chen2017stylebank, dumoulin2016learned,wang2017multimodal,li2017diversified, zhang2018multi} introduce {\color{black}{various}} network architectures to {\color{black}{simultaneously}} handle multiple styles; (3) \emph{Arbitrary-Style-Per-Model} methods \cite{huang2017arbitrary,li2017universal,sheng2018avatar,li2019learning,park2019arbitrary} further {\color{black}{adopt diverse}} feature modification mechanisms to transfer arbitrary styles.
Reviewing these methods, we find that although local style patterns can be transferred, complex style mixed {\color{black}{with}} both global and local patterns {\color{black}{is}} still not properly transferred. Meanwhile, artifacts and flaws {\color{black}{appear}} in many cases. 
{\color{black}{To this end}}, in this work, our main goal is to achieve superior high-quality artistic style transfer results with feed-forward network, where local and global patterns can be reserved aesthetically.

How human painters handle the complex style patterns {\color{black}{while}} painting?
A common process, especially for a beginner, is to first draw a draft to capture global structure and then revise the local details {\color{black}{gradually}}, instead of directly finishing the {\color{black}{final painting part-by-part}}.
Inspired by this, we propose a novel neural network named \textbf{Laplacian Pyramid Network (LapStyle)} for style transfer. 
%
Firstly, in our framework, a \emph{Drafting Network} is designed to transfer global style patterns in low-resolution, since we observe that global  patterns can be transferred easier in low resolution due to larger {\color{black}{receptive}} field and less local details.
{\color{black}{A \emph{Revision Network} is then used to revise the local details in high-resolution via hallucinating a residual image according to the draft and the textures extracted by Laplacian filtering over the 2$\times$ resolution content image. Note that our Revision Network can be stacked in a pyramid manner to generate higher resolution details. The final stylized image is obtained by aggregating outputs of all pyramid levels.}}
%
Further, we adopt shallow patch discriminators to adversarially learn local style patterns. 
As illustrated {\color{black}{in}} Fig. \ref{fig:overview}, {\color{black}{appealing}} stylization results are achieved by our ``Drafting and Revison'' process.
{\color{black}{To summarize}}, the main contributions are as follows:

\begin{itemize}
\item We introduce a novel framework ``Drafting and Revision'', which simulates painting creation mechanism by splitting style transfer process into global style pattern drafting and local style pattern revision.
\item We propose a novel feed-forward style transfer method named LapStyle. {\color{black}{It uses a Drafting Network to transfer global style patterns in low-resolution, and adopts higher resolution Revision Networks to revise local style patterns in a pyramid manner according to outputs of multi-level Laplacian filtering of the content image}}.
\item Experiments demonstrate that our method can generate high-resolution and high-quality stylization results, where global and local style patterns are both effectively synthesized. 
{\color{black}{Besides}}, the proposed LapStyle is extremely efficient {\color{black}{and}} can synthesize high resolution {\color{black}{stylized}} image of 512 pix in 110 fps.
\end{itemize}

\begin{figure*}[t]
\begin{center}
\begin{minipage}[b]{1.0\linewidth}
  \centering
  \centerline{\includegraphics[width=0.8\linewidth]{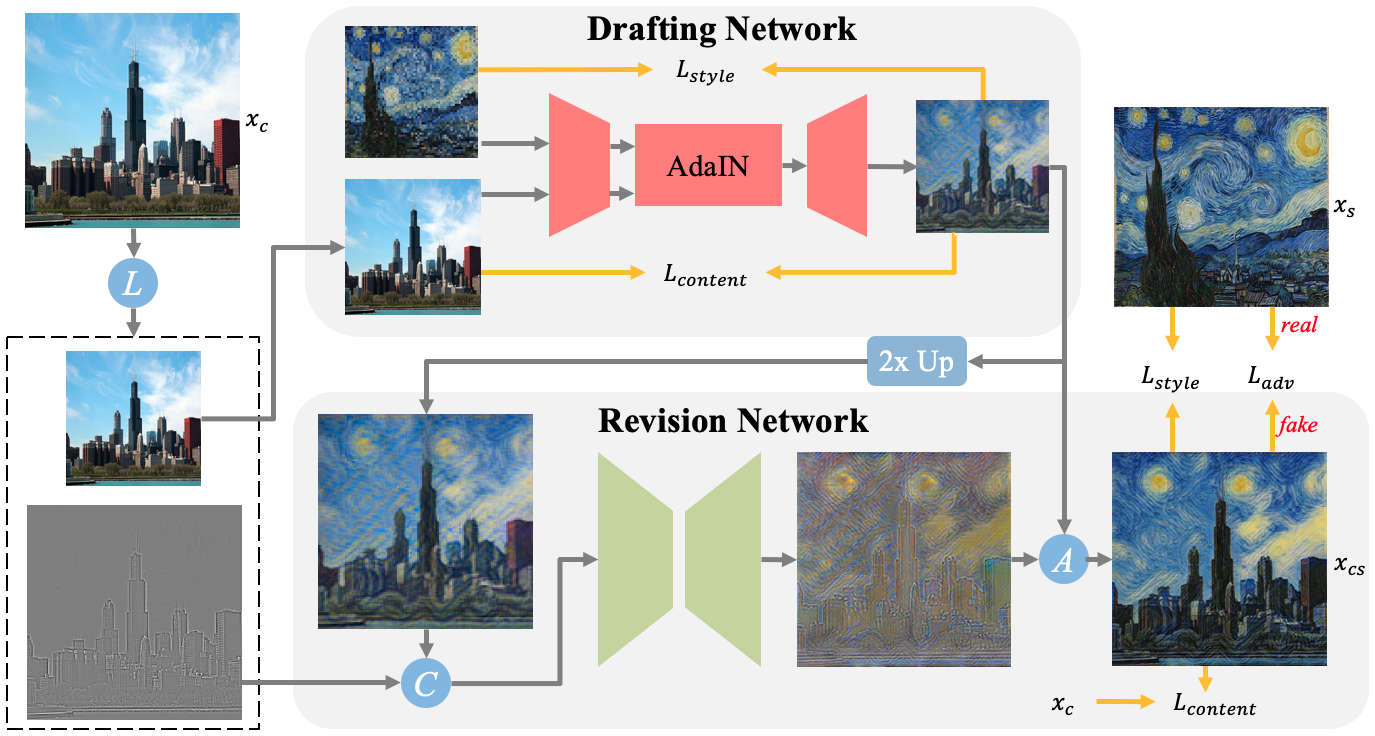}}
  \medskip
\end{minipage}
\end{center}
\caption{Overview of our framework. (a) We first generate {\color{black}{image pyramid $\{\bar{x}_c, r_c\}$ from content image $x_c$ with the help of Laplacian filter}}. (b) Rough low-resolution stylized image is generated by the Drafting Network. (c) Then the Revision Network generates stylized detail image in high resolution. (4) Final stylized image is generated by aggregating the outputs pyramid. \emph{L}, \emph{C} and \emph{A} in image represent Laplacian, concatenate and aggregation operation separately.}
\label{fig:framework}
\end{figure*}

\section{Related Work}

\noindent
\textbf{Style Transfer.}
 %
Style transfer algorithms aim at migrating the style from an example image to a content image. 
With the initiation of the seminal work Gatys \emph{et al.} \cite{gatys2016image}, various methods have been developed thereafter to address different aspects of, including visual quality \cite{li2016combining, gu2018arbitrary}, head portrait \cite{selim2016painting}, semantic control \cite{champandard2016semantic, gatys2017controlling} and so on.
Kolkin \emph{et al.} propose STROTSS \cite{kolkin2019style} and  higher quality stylized images can be generated by adopting Earth Movers Distance (rEMD) loss for optimization, which {\color{black}{deploys}} the style attributes with minimum distortion to content's semantic layout. 
However, the expensive computational cost of optimization-based methods hinder their practical applications.
%
In order to improve run-time efficiency, researchers have proposed to replace the iterative optimization procedure with feed-forward networks.  
\emph{Per-Style-Per-Model} methods \cite{johnson2016perceptual, li2016precomputed, ulyanov2016texture, ulyanov2016instance,ulyanov2017improved, chen2016fast} adopt auto-encoder as style transfer network trained with variants of content and style losses derived from \cite{gatys2016image}. \emph{Multi-Style-Per-Model} methods \cite{chen2017stylebank, dumoulin2016learned,wang2017multimodal,li2017diversified, zhang2018multi} embed learnable affine transformation architectures in the middle of auto-encoder to incorporate multiple styles. 
Recently, \emph{Arbitrary-Style-Per-Model} methods \cite{huang2017arbitrary,li2017universal,sheng2018avatar,li2019learning,park2019arbitrary} achieve arbitrary style transfer via style feature embedding networks. 
Besides, video style transfer methods \cite{chen2017coherent, huang2017real, chen2018stereoscopic} exploit how to generate stable stylization video.

Model-optimization based methods greatly improve computation efficiency with visual quality compromises. AdaIN \cite{huang2017arbitrary}, WCT \cite{li2017universal} and linear transformation \cite{li2019learning} adjust holistic feature distributions so they all fail to preserve local style patterns. SANet \cite{park2019arbitrary} embed local style patterns in content feature map with the aid of style attention mechanism, but they cannot perform well with large-scale textures such as the swirls in \emph{The Starry Night}. On the contrary, our proposed {\color{black}{LapStyle}} can capture the style statistics at different scales, which greatly improves the visual quality over current model-optimization based methods. 

\noindent
\textbf{Multi-scale Learning.}
%
In image manipulation area, working at multiple scales is a common technique to better capture a wide range of image statistics \cite{denton2015deep,lai2017deep,shaham2019singan,shih2014style,liao2017visual,wang2017multimodal,kolkin2019style}. Lai \emph{et al.} propose LapSRN \cite{lai2017deep} to progressively reconstruct the high-resolution images by predicting high-frequency residuals with cascaded convolutional networks. Shaham \emph{et al.} propose SinGAN \cite{shaham2019singan} to train the network with single image by capturing patch-level distribution at different image scales with a pyramid of adversarial networks. 
WCT \cite{li2017universal} and PhotoWCT \cite{li2018closed} also generate results coarse to fine gradually, but they work at the original RGB domain and not explicitly revise stylized details in the residual field as LapStyle does.
WCT$^2$ \cite{yoo2019photorealistic} also exploits residual information via wavelet transform where the residual information is mainly used to keep spatial details of original image. Differently, LapStyle constructs the Revision Network in the residual field to better transfer and enhance local stylization details. 
STROTSS \cite{kolkin2019style} also adopts a multi-scale scheme to apply style transfer by minimizing EMD loss at increasing resolution and exhibits high visual quality. 
However, the iterative optimization procedure suffers high computation cost and needs several minutes to synthesize one image. 
Our proposed {\color{black}{LapStyle}} captures a wide range of style statistics from global distribution to local patterns by adopting a multi-scale network to better balance the trade-off between run-time efficiency and visual quality.

\section{Approach}

In this section, we will introduce the proposed feed-forward style transfer network LapStyle in detail. 
{\color{black}{For ease of understanding, in this section, we only describe the framework with a 2-level pyramid. The base level is a Drafting Network and a Revision Network is used for the 2nd level of higher resolution, as shown in Fig. \ref{fig:framework}. It is quite straightforward to build more levels by stacking Revision Networks.}} 
%

\subsection{Network Architecture}

Our proposed LapStyle takes a content image $x_c\in R^{H_c\times W_c}$ and a pre-defined style image $x_s$ as inputs, and eventually synthesizes a stylized image $x_{cs}$. 
%
%
%
As shown in Fig. \ref{fig:framework}, for pre-processing, {\color{black}{we construct a 2-level image pyramid $\{\bar{x}_c, r_c\}$. $\bar{x}_c$ is simply a 2$\times$ downsampled version of $x_c$. $r_c$ is obtained with the help of Laplacian filter, \ie, $r_c = x_c - Up(\mathcal{L}(\bar{x}_c))$, where $\mathcal{L}$ denotes Laplacian filtering and $Up$ is 2$\times$ upsample operation. The style image $x_s$ is also downsampled to a low-resolution version $\bar{x}_s$. 
}}  
%

In the {first stage}, the \emph{Drafting Network} first {\color{black}{encodes context and style features}} from both $\bar{x}_c$ and $\bar{x}_s$ with a pre-trained neural network, then it modulates content feature using style feature in multiple granularities and finally generates the stylized image $\bar{x}_{cs} \in R^{H_c/2 \cdot W_c/2}$ using a decoder.
In the {second stage}, the \emph{Revision Network} first up-samples $\bar{x}_{cs}$ to ${x}_{cs}'\in R^{H_c \cdot W_c}$, then it concatenates ${x}_{cs}'$ and $r_c$ as the input to generate stylized a detail image $r_{cs} \in R^{H_c \cdot W_c}$.
Finally, we obtain stylized image $x_{cs} \in R^{H_c \cdot W_c}$ by aggregating the pyramid outputs:

\begin{equation}
x_{cs} = \mathcal{A}(\bar{x}_{cs}, r_{cs}),
\end{equation}
where $\mathcal{A}$ denotes the aggregation function. In the following, we will introduce the configuration of Drafting Network and Revision Network in detail.

\begin{figure}[t]
\begin{center}
\begin{minipage}[b]{1.0\linewidth}
  \centering
  \centerline{\includegraphics[width=8.5cm]{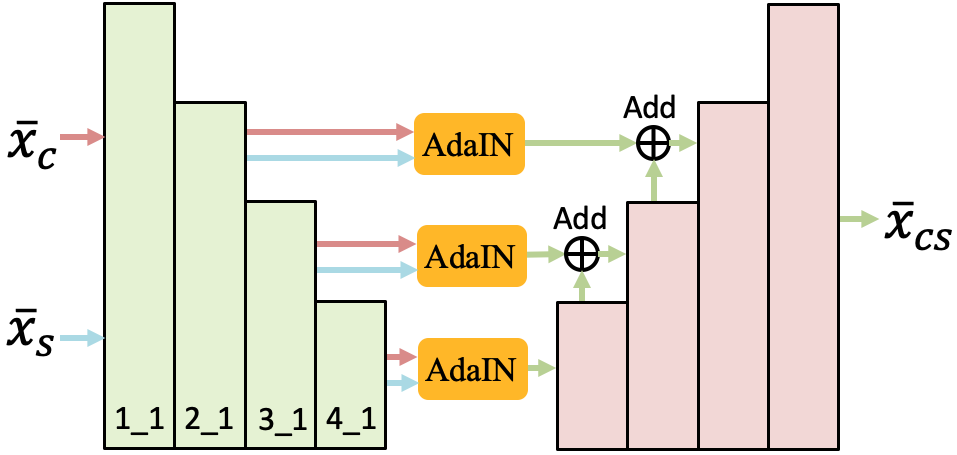}}
  \medskip
\end{minipage}
\end{center}
\caption{Illustration of the proposed Drafting Network.}
\label{fig:draftnet}
\end{figure}

\subsection{Drafting Network}

The Drafting Network aims at synthesizing  global style patterns in low resolution. Why low resolution? As demonstrated in Section \ref{ablation}, we observe that global  patterns can be transferred easier in low resolution, due to large receptive field and less local details.
To achieve single style transfer, earlier work \cite{johnson2016perceptual} directly trains an encoder-decoder module, where only content image is used as input.
%
To better combine the style feature and the content feature, we adopt AdaIN module from recent arbitrary style transfer method \cite{huang2017arbitrary}. 

The architecture of Drafting Network is shown in Fig \ref{fig:draftnet}, which includes an encoder,  several AdaIN modules and a decoder.
\textbf{(1)} The encoder is a pre-trained VGG-19 network, which is fixed during training. Given $\bar{x}_c$ and $\bar{x}_s$, the VGG encoder extracts features in multiple granularity at $2\_1$,  $3\_1$ and $4\_1$ layers.
%
\textbf{(2)} Then, we apply feature {\color{black}{modulation}} between the content and style feature using AdaIN modules after $2\_1$, $3\_1$ and $4\_1$ layers, respectively.
%
\textbf{(3)} Finally, in each granularity of decoder, {\color{black}{the corresponding feature from the AdaIN module is merged via a skip-connection.
Here, skip-connections after AdaIN modules in both low and high levels are leveraged}} to help to reserve content structure, especially for low-resolution image.



\begin{figure}[t]
\begin{center}
\begin{minipage}[b]{1.0\linewidth}
  \centering
  \centerline{\includegraphics[width=8.0cm]{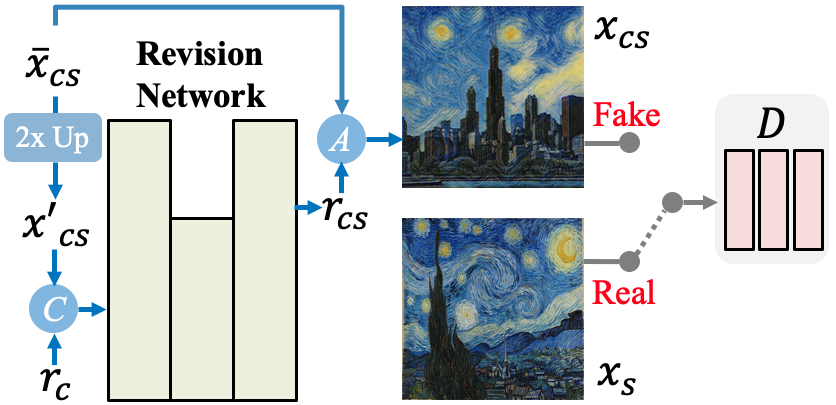}}
  \medskip
\end{minipage}
\end{center}
\caption{Illustration of the proposed Revision Network. $C$ and $A$ here represent concatenate and aggregation operation separately.}
\label{fig:revnet}
\end{figure}

\subsection{Revision Network}

The Revision Network aims to revise the rough stylized image  via generating {\color{black}{residual details}} image $r_{cs}$, while the final stylized image is generated by combining $r_{cs}$ and rough stylized image $\bar{x}_{cs}$. This procedure ensures that the distribution of global style pattern in $\bar{x}_{cs}$ is properly kept. Meanwhile, learning to revise local style patterns with residual details image is easier for the Revision Network.

As shown in Fig. \ref{fig:revnet}, the Revision Network {\color{black}{is designed as}} a simple yet effective encoder-decoder architecture, with only one down-sampling and one up-sampling layer.
Further, we introduce a patch discriminator to help Revision Network to {\color{black}{capture}} fine patch textures under adversarial learning setting. We define the patch discriminator $D$ following SinGAN \cite{shaham2019singan}, where $D$ owns 5 convolution layers and 32 hidden channels. We choose to define a relatively shallow $D$ to (1) avoid over-fitting since we only have one style image and (2) control the receptive field to ensure $D$ can only capture local patterns.

\subsection{Training}

During training, the Drafting Network and the Revision Network are both optimized with content and style loss, while the Revision Network further adopts adversarial loss.
Thus, we first describe style and content losses, then introduce the full objective of two networks separately. Since our LapStyle is ``Per-Style-Per-Model'', during training, we keep a single $x_s$, and a set of $x_c$ from content dataset $X_c$.

\noindent
\textbf{Style Loss.}
Following recent optimization-based method STROTSS \cite{kolkin2019style}, we {\color{black}{combine}} the relaxed Earth Mover Distance (rEMD) loss and {\color{black}{the}} commonly used mean-variance loss as style loss. 
To begin with, given an image, we can use pre-trained VGG-19 encoder to extract a set of feature vectors as $\mathcal{F} = \{F^{1\_1}, F^{2\_1}, F^{3\_1}, F^{4\_1}, F^{5\_1}\}$.
%
%
The rEMD loss aims at {\color{black}{measuring}} the distance between the feature distributions of style image $x_s$ and stylized image $x_{cs}$. {\color{black}{For simplicity, we omit the superscripts which indicate layer index in the following. Supposing $F_s\in R^{h_sw_s\times c}, F_{cs}\in R^{h_{cs}w_{cs}\times c}$ are the features of $x_s$ and $x_{cs}$,
their rEMD loss can be calculated as:

\begin{equation}
l_{r}  = \max\left (\frac{1}{h_sw_s}\sum_{i=1}^{h_sw_s} \underset{j}{\min}C_{ij}, \frac{1}{h_{cs}w_{cs}}\sum_{j=1}^{h_{cs}w_{cs }} \underset{i}{\min}C_{ij}\right ),
\end{equation}
where the cosine distance term $C_{ij}$ is defined as:

\begin{equation}
C_{ij} = 1 - \frac{F_{s,i} \cdot F_{cs,j}}{\left \| F_{s,i} \right \| \left \| F_{cs,j} \right \|}
\end{equation}
}}
To keep the magnitude of the feature vectors, we also adopt the commonly used  mean-variance loss as:

\begin{equation}
l_m = \left \| \mu (F_s) - \mu (F_{cs})  \right \| _2 + \left \| \sigma (F_s) - \sigma (F_{cs})  \right \|_2,
\end{equation}
where $\mu$ and $\sigma$ calculate the mean and co-variance of the feature vectors separately.

\noindent
\textbf{Content Loss.}
For content loss, we adopt the commonly used normalized perceptual loss and the self similarity loss {\color{black}{between $F_c\in R^{h_cw_c\times c}$ and $F_{cs}\in R^{h_{cs}w_{cs}\times c}$ proposed in \cite{kolkin2019style}. Note that $h_{cs}$ equals $h_c$ and $w_c$ equals $w_{cs}$ because of $x_c$ and $x_cs$ are of the same resolution.
The perceptual loss is defined as:
\begin{equation}
l_p =  \left \| norm(F_c) - norm(F_{cs})  \right \| _2,
\end{equation}
where $norm$ denotes the channel-wise normalization for $F$.
The self-similarity loss aims to retain the relative relation in content image to stylized image, which is defined as:

\begin{equation}
l_{ss} = \frac{1}{(h_cw_c)^2} \sum_{i,j} \left |\frac{D_{ij}^c}{\sum_i D_{ij}^c} - \frac{D_{ij}^{cs}}{\sum_i D_{ij}^{cs}}  \right |,
\end{equation}
here $D_{ij}^c$ and $D_{ij}^{cs}$ are the $(i,j)^{th}$ entry of self-similarity matrices $D^c$ and $D^{cs}$, respectively. Here $D_{ij}$ is pairwise cosine similarity $<F_i, F_j>$.
}}

\noindent
\textbf{Loss of Drafting Network.}
In training phase of Drafting Network, low resolution images $\bar{x}_c$ and $\bar{x}_s$ are used as the network input, and also are used to calculate the content loss and style loss separately.
The overall training objective function of the Drafting Network is defined as:

\begin{equation}
L_{Draft}  = {(l_p + \lambda_{1} \cdot l_{ss})+ \alpha \cdot (l_m + \lambda_{2} \cdot l_r )}
\end{equation}
where $\lambda_{1}$, $\lambda_{2}$ and $\alpha$ are weight terms. We control the balance of content and style loss via adjusting $\alpha$. Specifically, $l_r$ and $l_{ss}$ are defined on $3\_1$ and $4\_1$ layers, meanwhile $l_m$ and $l_p$ are defined from $1\_1$ to $5\_1$ layers.

\noindent
\textbf{Loss of Revision Network.}
In training phase of Revision Network, the parameters of Drafting Network are fixed and the training loss is built upon $x_{cs}$. 
To better learn local fine-grain textures, except for base content and style loss $L_{base} = L_{Draft}$, we introduce a discriminator and train Revision Network with an adversarial loss term. The overall optimization objective is defined as:

\begin{equation}
 \underset{Rev}{\min}L_{base} + \beta \cdot \underset{Rev}{\min}\  \underset{D}{\max} L_{adv}(Rev, D),
\end{equation}
where $Rev$ denotes the Revision Network, $D$ denotes the discriminator, and $\beta$ controls the balance between base style transfer losses and adversarial loss. {\color{black}{$L_{adv}$ is standard adversarial training loss.}}

\section{Results}

\subsection{Dataset and Setup}

\noindent
\textbf{Dataset.}
To train our model, we need a single style image and a collection of content images. In this work, following conventions, we use MS-COCO \cite{lin2014microsoft} as content images and select style images from WikiArt \cite{phillips2011wiki}. Some copyright free images from \emph{Pexels.com} is also used as style images.

\begin{figure}[t]
\begin{center}
\begin{minipage}[b]{1.0\linewidth}
  \centering
  \centerline{\includegraphics[width=6.0cm]{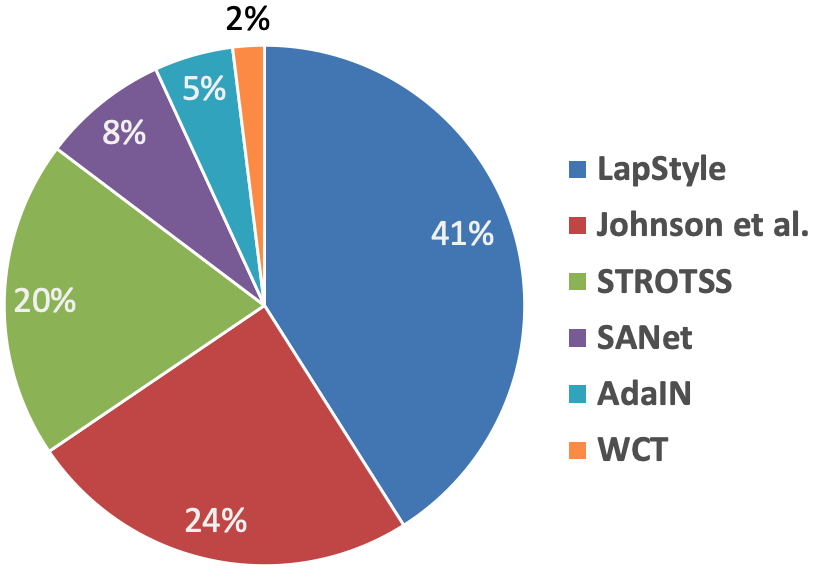}}
  \medskip
\end{minipage}
\end{center}
\caption{User preference results of five SOTA methods.}
\label{fig:exp_user}
\end{figure}

\begin{table}
\begin{center}
    \begin{tabular}{ccc}
    \toprule
    Method & Time (256pix) & Time(512pix) \\
    \hline
    Gatys et al. \cite{gatys2016image} & 15.863 & 50.804 \\
    STROTSS \cite{kolkin2019style} & 163.052 & 177.485 \\
    Johnson et al. \cite{johnson2016perceptual} & 0.132 & 0.149 \\
    WCT \cite{li2017universal} & 0.689 & 0.997 \\
    AdaIN \cite{huang2017arbitrary} & 0.011 & 0.039 \\
    Linear \cite{li2019learning} & 0.007 & 0.039 \\
    SANet \cite{park2019arbitrary} & 0.017 & 0.055 \\
    \hline
    Ours & 0.008 & 0.009\\
    \bottomrule
    \end{tabular}
\end{center}

\caption{Execution time comparison (in seconds).}
\label{table:speed}
\end{table}

\noindent
\textbf{Implementation Details.}
In LapStyle, Drafting and Revision Networks are trained in sequence. The former one is first trained with resolution of $128\times128$, then the later one is trained with $256\times256$. To achieve higher resolution, we can consecutively train more Revision Networks using resolution of 512 and 1024.
For both networks, we use the Adam optimizer \cite{kingma2014adam} with a learning rate of 1e-4 and a batch size of 5 content images. For both networks, the training process consists of 30,000 iterations.
The loss weight term, $\lambda _1$, $\lambda _2$, $\alpha$ and $\beta$ are set to 16, 3, 3 and 1, respectively.
More detailed network configurations of LapStyle is {\color{black}{presented in our supplementary material}}. 

\begin{figure*}[htb]
\begin{center}
\begin{minipage}[b]{1.0\linewidth}
  \centering
  \centerline{\includegraphics[width=16.1cm]{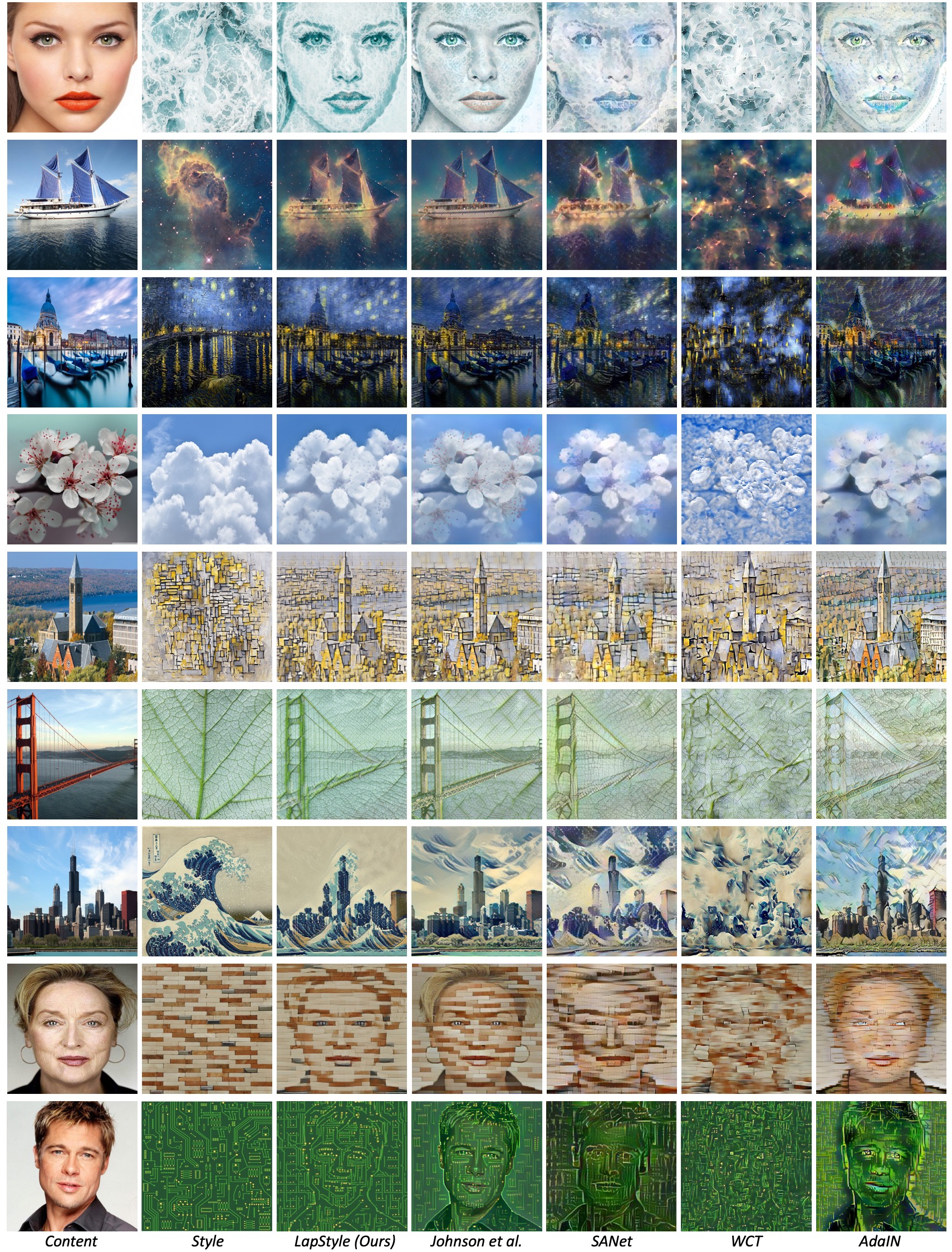}}
  \medskip
\end{minipage}
\end{center}
\caption{Qualitative comparisons between our method and state-of-the-art feed forward methods.}
\label{fig:exp_full_page}
\end{figure*}

\begin{figure*}[th]
\begin{center}
\begin{minipage}[b]{1.0\linewidth}
  \centering
  \centerline{\includegraphics[width=17.0cm]{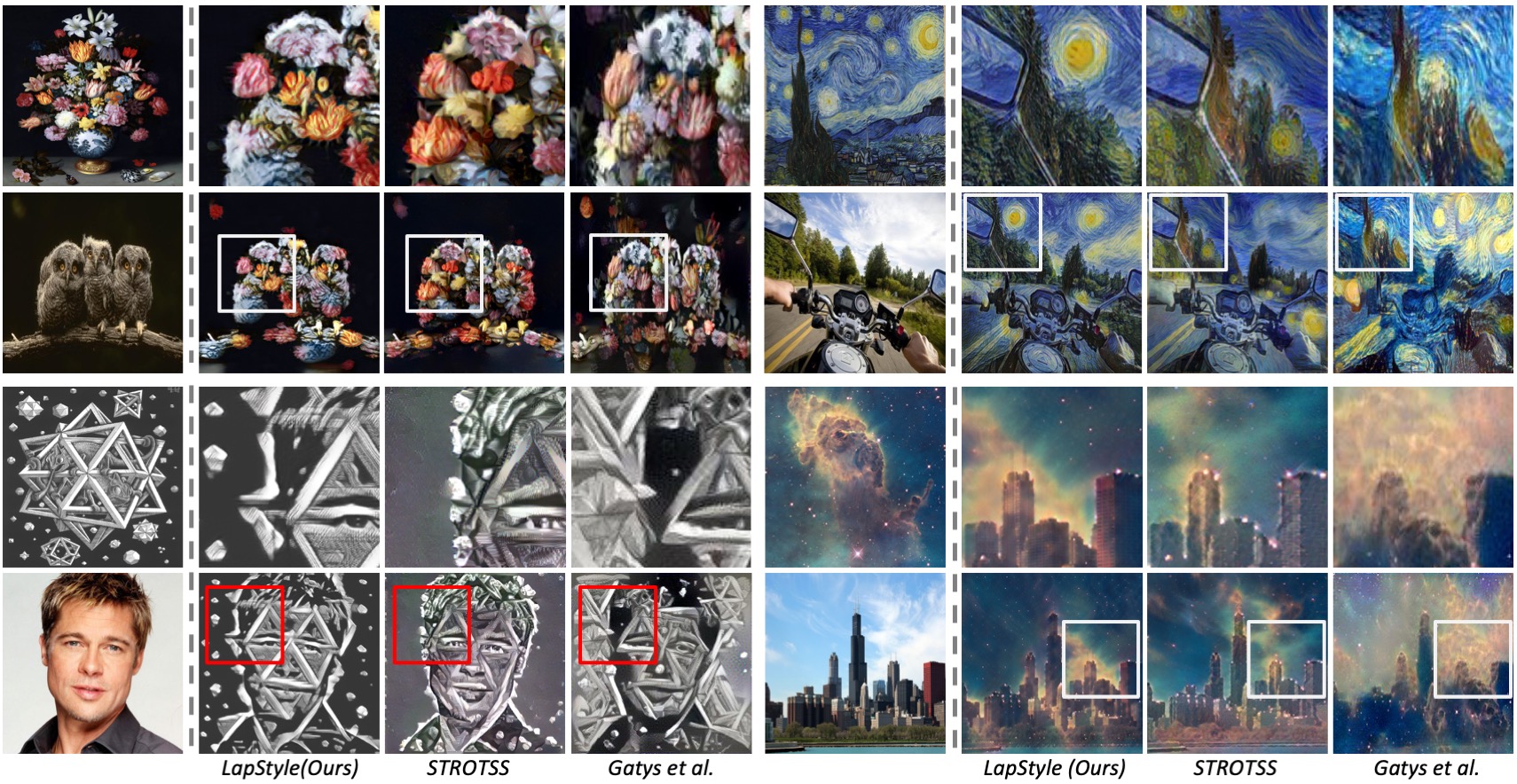}}
  \medskip
\end{minipage}
\end{center}
\caption{Qualitative comparisons between our method and optimization-based methods STROTSS \cite{kolkin2019style} and Gatys et al. \cite{gatys2016image}.}
\label{fig:exp_vis_opt}
\end{figure*}

\subsection{Comparison with Prior Works}

\noindent
\textbf{Qualitative Comparison.}
As shown in Fig. \ref{fig:exp_full_page}, we compare our method with state-of-the-art feed-forward methods.
%
Like our LapStyle, Johnson et al. \cite{johnson2016perceptual} is also a single style transfer method. \cite{johnson2016perceptual} can synthesize some local style patterns with clear structure (e.g. row 8), however, the color distributions and texture structures of content image are often maintained (e.g. rows 2 and 8), resulting in insufficient stylization. 
AdaIN \cite{huang2017arbitrary}, WCT \cite{li2017universal} and SANet \cite{park2019arbitrary} are arbitrary style transfer models, which have some common features: (1) they mainly transfer the color distribution and simple local patterns of style image; (2) the complex style patterns with bigger size is basically not transferred (e.g. rows 2, 5 and 6); (3) the local patterns are usually not accurately transferred, resulting in messy local textures (e.g. rows 2, 8).
To be specific, the problem of retaining color distribution of content is severe in AdaIN \cite{huang2017arbitrary} (e.g. rows 2, 7 and 8). WCT \cite{li2017universal} discards too much context structure, resulting in disordered and chaotic image.
In contrast to these methods,  our method can simultaneously transfer simple local style patterns and complex global style patterns, retaining clear and clean structure of style patterns. The color distribution is also completely transferred.
Although LapStyle can not transfer arbitrary style, we believe that improving the stylization quality is most important for feed-forward method. We leave arbitrary LapStyle for future work.

In Fig. \ref{fig:exp_vis_opt}, we show some stylization examples synthesized by our method and two optimization-based style transfer methods \cite{kolkin2019style, gatys2016image}, where zoom-in views are also demonstrated for better comparison.
Gatys et al. \cite{gatys2016image} synthesis stylized image in single scale via optimizing gram matrix. As shown in Fig. \ref{fig:exp_vis_opt}, although holistic style patterns are transferred, the distribution of style patterns are often inappropriate (e.g. left-down and right-down). Meanwhile, the color distribution of stylized image is not accurate enough.
STROTSS \cite{kolkin2019style} is the state-of-the-art optimization-based method, which synthesizes stylized image in multiple scales sequentially (from 32 pix to 512 pix) with EMD loss. As shown in Fig.  \ref{fig:exp_vis_opt}, the stylized images have delicate texture and clear style pattern.
As a feed-forward method, our method achieves comparable stylization results with STROTSS. In some cases (e.g. top-right and bottom-left in Fig. \ref{fig:exp_vis_opt}), large scale patterns are better synthesized by our method. The comparative advantage of STROTSS is that style patterns and context structures are combined better in some cases (e.g. bottom-right in Fig. \ref{fig:exp_vis_opt}), due to its optimization process.


\noindent
\textbf{User Study.}
We choose 15 style images and 15 content images to synthesize 225 images in total using our method and 5 {\color{black}{competitive}} SOTA methods.
Then, we randomly sample 20 content-style pairs. 
For each pair, we display stylized images side-by-side in a random order to subjects and ask them to select their most favorite one. As shown in Fig \ref{fig:exp_user}, we collect 2000 votes from 100 users and show the percentage of votes for each method in the form of pie chart.
The comparison results demonstrate that our stylized results are significantly {\color{black}{more appealing than competitors}}.

\begin{figure*}[t]
\begin{center}
\begin{minipage}[b]{1.0\linewidth}
  \centering
  \centerline{\includegraphics[width=16.75cm]{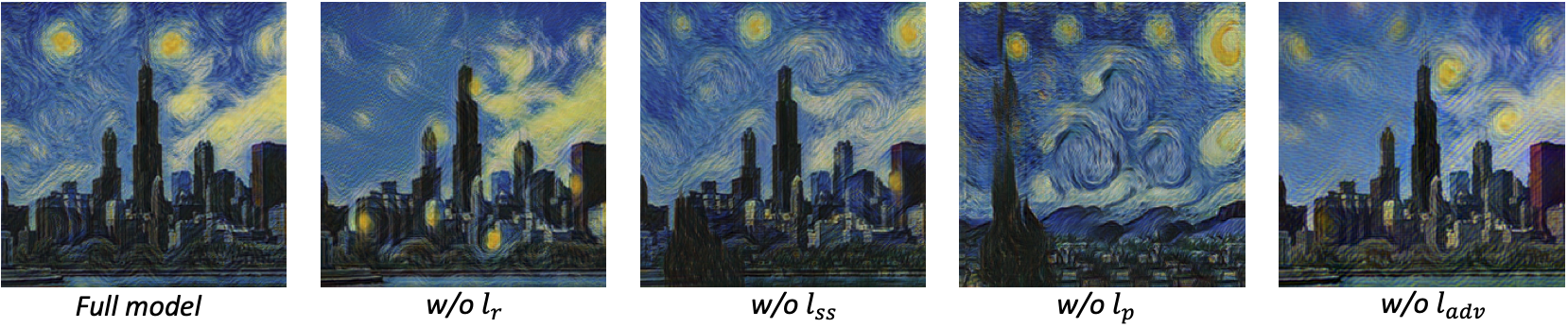}}
  \medskip
\end{minipage}
\end{center}
\vspace{-0.3cm}
\caption{Ablation study of effects of loss function used during training. Here, $l_r$, $l_{ss}$ and $l_p$ are used in both networks, while $l_{adv}$ only used in the Revision Network. $l_m$ is used in all ablation settings to keep style transferred.  Best viewed zoomed-in on screen.}

\label{fig:exp_ab_loss}
\end{figure*}

\noindent
\textbf{Efficiency Analysis.}
We compare the efficiency of our proposed method and other optimization methods and feed-forward methods. Two image scales are used during evaluation: 256 and 512 resolution. For 512-pixel inference, two Revision Networks are used.
All experiments are conduct using a single Nvidia Titan X GPU.
The comparison results are shown in Table. \ref{table:speed}, where we can find our method runs in real-time and achieves 120 fps and 110 fps with 256-pix and 512-pix, respectively. 
There are three reasons for the fast inference speed: (1) The Drafting Network is build upon low-resolution; (2) AdaIN module is efficient and (3) the Revision Network is shallow.
As shown in Fig. \ref{fig:exp_full_page} and Fig. \ref{fig:exp_vis_opt}, the quality of  stylized image generated by our methods is comparable with optimization-based method, and is significantly better than feed-forward methods.
To conclude, Table. \ref{table:speed} demonstrates that our method achieves the SOTA inference speed among feed-forward methods, and is much more faster than optimization methods.


%

\subsection{Ablation Study}
\label{ablation}

\noindent
\textbf{Loss Function.}
We conduct ablation experiments to verify the effectiveness of each loss term used for training LapStyle, the results are shown in Fig. \ref{fig:exp_ab_loss}.
(1) Without rEMD loss $l_r$, the {\color{black}{style patterns of yellow circle}} disappear and the overall stylization degree is decreased. This result demonstrates the effectiveness of rEMD loss, and we are the first to train feed-forward network with rEMD loss;
(2) Without self-similarity loss $l_{ss}$, some inappropriate black style patterns appear in the bottom-left corner. 
(3) Without perceptual loss $l_{p}$, the structure of content image is totally discarded and the LapStyle directly re-builds the style image. These results suggest that $l_p$ is necessary for our method, meanwhile $l_{ss}$ can further constrain the content consistency to achieve better style distribution.
(4) Without adversarial loss $l_{adv}$, the texture quality and color distribution become worse than full model. This comparison demonstrates that the adversarial learning in revision phase can effectively improve the stylization quality, especially local texture and color distribution.

\begin{figure}[t]
\begin{center}
\begin{minipage}[b]{1.0\linewidth}
  \centering
  \centerline{\includegraphics[width=8.0cm]{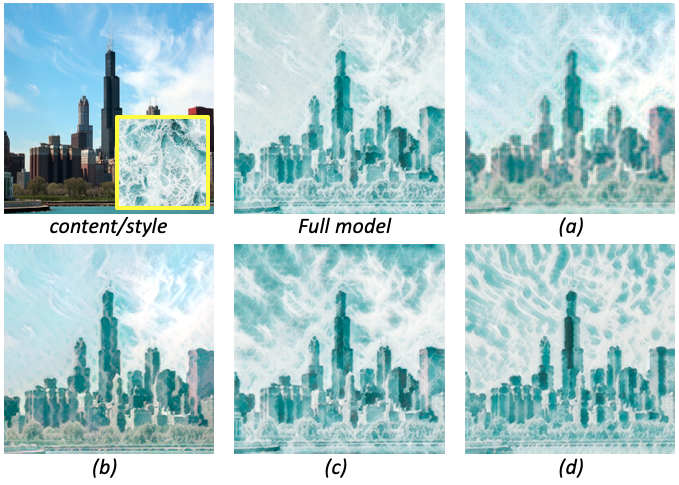}}
  \medskip
\end{minipage}
\end{center}
\caption{Ablation study of effectiveness of the Revision Network. (a) Drafting Network trained on 128 pix (result image is up-sampled to 256 pix). (b) Drafting Network directly trained on 256 pix. In (a) and (b), Revision Network is not used. (c) Revision Network is directly built upon RGB image instead of difference image. (d) Revision Network is trained without Drafting Network.}
\label{fig:exp_ab_net}
\end{figure}

\noindent
\textbf{Effectiveness of Revision Network.}
The results of ablation experiments are shown in Fig. \ref{fig:exp_ab_net}.
Before revision, the result of Drafting Network is blur in Fig. \ref{fig:exp_ab_net} (a), due to low resolution. If we directly train Drafting Network on 256 pix, as Fig. \ref{fig:exp_ab_net} (b) shows, the result is clear but its stylization degree is limited. These results demonstrate the effectiveness and necessary of our ``Drafting and Revison'' framework. 
{\color{black}{Another question is whether it is necessary to revise the rough stylized image with the help of Laplacian difference image? The image of Fig. \ref{fig:exp_ab_net} (c) is directly generated by the Revision network in RGB space}}. We can see the style distribution of revision result is divorced from the drafting image (a) and it seems less harmony than the original result. This observation suggests revising stylized image in a residual form is more controllable and can generate better results.

\noindent
\textbf{Effectiveness of Drafting Network.} As shown in Fig. \ref{fig:exp_ab_net}, without DraNet, the RevNet can still capture style patterns to some extend, but significantly worse than full model.

\begin{figure}[t]
\begin{center}
\begin{minipage}[b]{1.0\linewidth}
  \centering
  \centerline{\includegraphics[width=8.0cm]{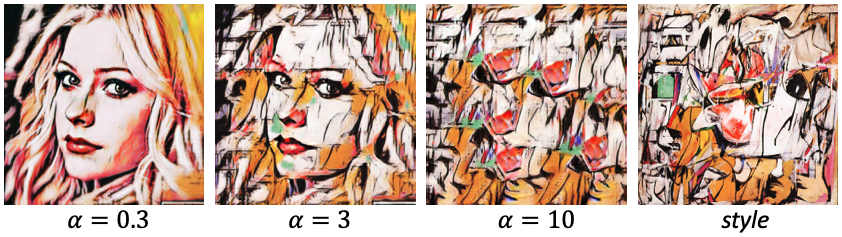}}
  \medskip
\end{minipage}
\end{center}
\caption{Trade-off of content-style losses. }
\label{fig:exp_ab_content_style}
\end{figure}

\noindent
\textbf{Content-style Tradeoff.}
In the training phase, we can control the stylization degree by adjusting the weight term $\alpha$. As shown in Fig. \ref{fig:exp_ab_content_style}, the network tends to preserve more details and structures of the content image with low style loss, and synthesize excess style patterns with high style loss.


\section{Conclusion}

In conclusion, we propose a new feed-forward style transfer algorithm LapStyle which synthesizes stylized image in a progressive procedure.
In LapStyle, we propose the novel framework ``Drafting and Revision'', which first synthesizes a rough drafting with global pattern and then revises local style patterns {\color{black}{according to residual image generated with the help of Laplacian filtering}}.
Experiments demonstrate that our method is effective and efficient. It can synthesize images that are preferred over other state-of-the-art feed-forward style transfer algorithms and can run in real-time. {\color{black}{Currently, our LapStyle is designed following the \emph{Per-Style-Per-Model} fashion, arbitrary style transfer is left to be our future work}}. 
%
%

{\small
	\bibliographystyle{ieee_fullname}
	\bibliography{egbib}
}

\newpage

\newpage

\section{Supplementary}

The {\color{black}{outline}} of this supplementary material is as follows:
\begin{itemize}
\item The architecture details of our proposed LapStyle.
\item {\color{black}{A}} video stylization demo generated by LapStyle.
\item Qualitative results of high resolution stylization. 
\end{itemize}

\subsection{Network Structure}
In this supplementary material, we introduce the detailed configuration of our proposed LapStyle, which is comprised of a Drafting Network and a Revision Network.

The Drafting Network contains an encoder and a decoder. The network configuration of encoder is shown in Table. \ref{tab:t1}, which is a part of VGG-16 network and pre-trained on ImageNet dataset. We do not optimize the encoder during training. The architecture of decoder is shown in Table. \ref{tab:t2}, where AdaIN modules are added in multiple granularity to ensure style patterns of different granularity are properly transferred.
The Revision Network (Table. \ref{tab:t3}) has a simple encoder-decoder architecture of only 6 convolution layers. The light architecture ensures fast speed of the Revision Network, meanwhile restricts the potential of Revision Network so that it can only handle local style patterns.


\begin{table}[t]
\centering\noindent
\begin{center}
    \caption{The encoder architecture of the Drafting Network. The shape of an input sample is $3\times128\times128$. We use ``same padding'' for each convolution layer, and ReLU is added to the end of each convolution layer.}%
\label{tab:t1}
\begin{tabular}{c|c|c}
\hline
 stage            & output                                 &  architecture  \\ \hline
$F_{1\_1}$    & $64\times 128 \times 128$  &   \begin{tabular}{c} $1\times1$  conv, 3  \\
$3\times3$  conv, 64 \end{tabular}  \\ \hline
$F_{1\_2}$    & $64\times 128 \times 128$  &  $3\times3$  conv, 64  \\ \hline
             & $64\times 64 \times 64$  &  $2\times2$  max pooling, stride 2  \\ \hline
$F_{2\_1}$    & $128\times 64 \times 64$  &  $3\times3$  conv, 128 \\ \hline
$F_{2\_2}$    & $128\times 64 \times 64$  &  $3\times3$  conv, 128 \\ \hline
             & $128\times 32 \times 32$  &  $2\times2$  max pooling, stride 2  \\ \hline
$F_{3\_1}$    & $256\times 32 \times 32$  &  $3\times3$  conv, 256 \\ \hline
$F_{3\_2}$    & $256\times 32 \times 32$  &  $3\times3$  conv, 256 \\ \hline
             & $256\times 16 \times 16$  &  $2\times2$  max pooling, stride 2  \\ \hline
$F_{4\_1}$    & $512\times 16 \times 16$  &  $3\times3$  conv, 512 \\ \hline
\end{tabular}
\end{center}
\end{table}

\begin{table}[t]
\centering\noindent
\begin{center}
    \caption{The decoder architecture of the Drafting Network. $AdaIN$ is used in $4\_1$, $3\_1$, $2\_1$. A resblock consists of a $3\times 3$ convolution layer with $ReLU$ activation, an $1\times 1$ convolution layer and a residual connection. }
\label{tab:t2}
\begin{tabular}{c|c|c}
\hline
 stage            & output            &   architecture \\ \hline
$F^d_4$    & $256\times 16 \times 16$  &   \begin{tabular}{c} AdaIN($F_{4\_1}^c$, $F_{4\_1}^s$) \\
ResBlock(512) \\
$3\times3$ conv, 256, $Relu$ \end{tabular} \\ \hline
$F^d_3$     & $128\times 32 \times 32$  &   \begin{tabular}{c} upsample, scale 2 \\
add AdaIN($F_{3\_1}^c$, $F_{3\_1}^s$) \\
ResBlock(256) \\
$3\times3$ conv, 128, $Relu$ \end{tabular} \\ \hline
$F^d_2$     & $64\times 64 \times 64$  &   \begin{tabular}{c} upsample, scale 2 \\
add AdaIN($F_{2\_1}^c$, $F_{2\_1}^s$) \\
$3\times3$ conv, 128, $Relu$ \\
$3\times3$ conv, 64, $Relu$ \end{tabular} \\ \hline
$F^d_1$     & $3\times 128 \times 128$  &   \begin{tabular}{c} upsample, scale 2 \\
$3\times3$ conv, 64, $Relu$ \\
$3\times3$ conv, 3 \end{tabular} \\ \hline
\end{tabular}
\end{center}
\end{table}

\begin{table}[t]
\centering\noindent
\begin{center}
    \caption{The architecture of the Revision Network. The shape of input sample is $6\times H \times W$, which is the concatenated tensor combined by different image and stylized image generated by the Drafting Network (or the Revision Network at the previous scale). }
\label{tab:t3}
\begin{tabular}{c|c|c}
\hline
 stage            & output            &   architecture \\ \hline
$R_1$    & $64\times \frac{H}{2} \times \frac{W}{2} $  &   \begin{tabular}{c} $3\times3$ conv, stride 1, 64, $Relu$ \\
$3\times3$ conv, stride 2, 64, $Relu$ \end{tabular} \\ \hline
$R_2$    & $64\times \frac{H}{2} \times \frac{W}{2}$  &  ResBlock(64) \\ \hline
$R_3$     & $3\times H \times W$  &   \begin{tabular}{c} upsample, scale 2 \\
$3\times3$ conv, stride 1, 64, $Relu$ \\
$3\times3$ conv, stride 1, 3\end{tabular} \\ \hline
\end{tabular}
\end{center}
\end{table}

\subsection{Video Stylization}

\begin{figure*}[htb]
\begin{center}
\begin{minipage}[b]{1.0\linewidth}
  \centering
  \centerline{\includegraphics[width=15cm]{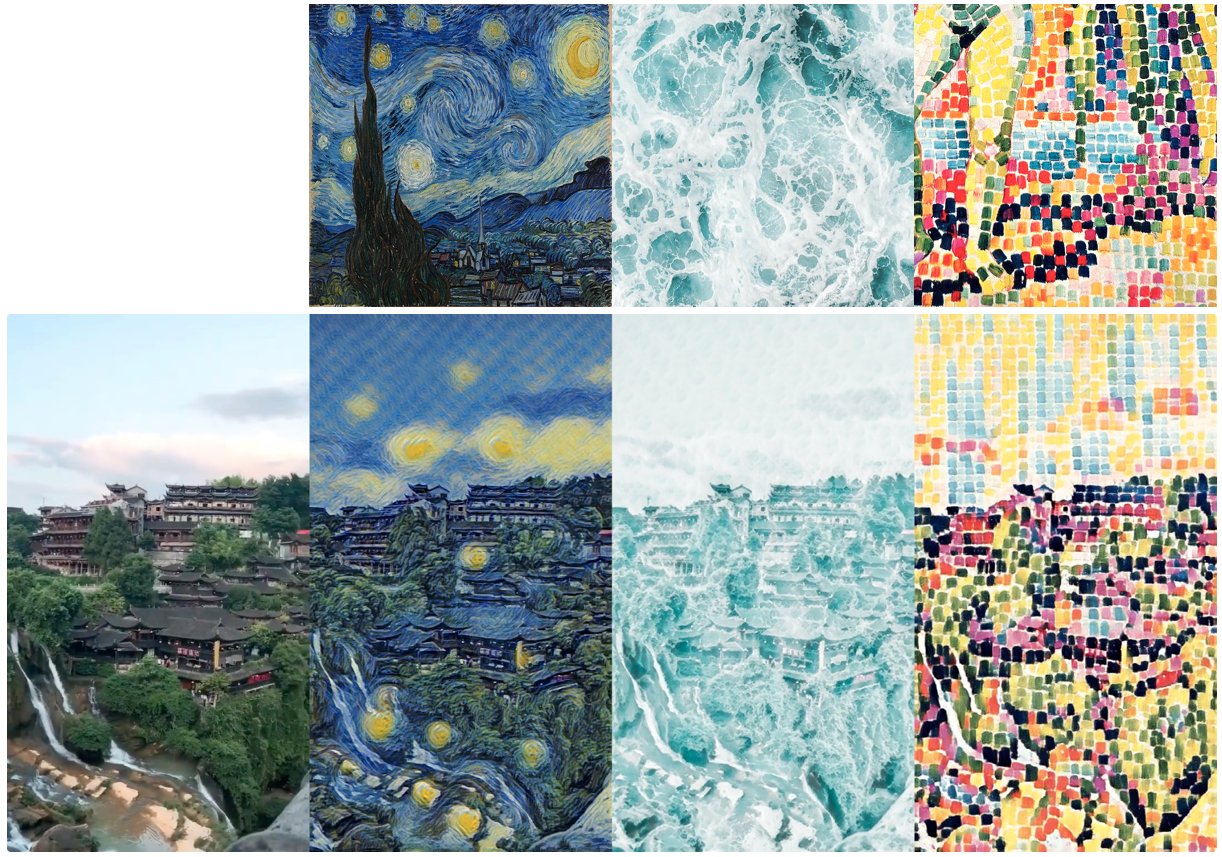}}
  \medskip
\end{minipage}
\end{center}
\caption{Video stylization demo. This figure illustrates a single frame from video, which can be found in supplementary material.}
\label{fig:exp_video}
\end{figure*}

As stated in the part of efficiency analysis in our main draft, our LapStyle can synthesis stylized image in real-time. Thus, it is suitable for video stylization application. In Fig. \ref{fig:exp_video}, we demonstrate the style image and one frame of video stylization demo. 
As shown in the video, the synthesized style patterns are generally stable with jitter to some extent. These jitter can be further removed by optical-flow based smoothness.

%
%
 
\subsection{High Resolution Stylization}

\begin{figure*}[htb]
\begin{center}
\begin{minipage}[b]{1.0\linewidth}
  \centering
  \centerline{\includegraphics[width=16cm]{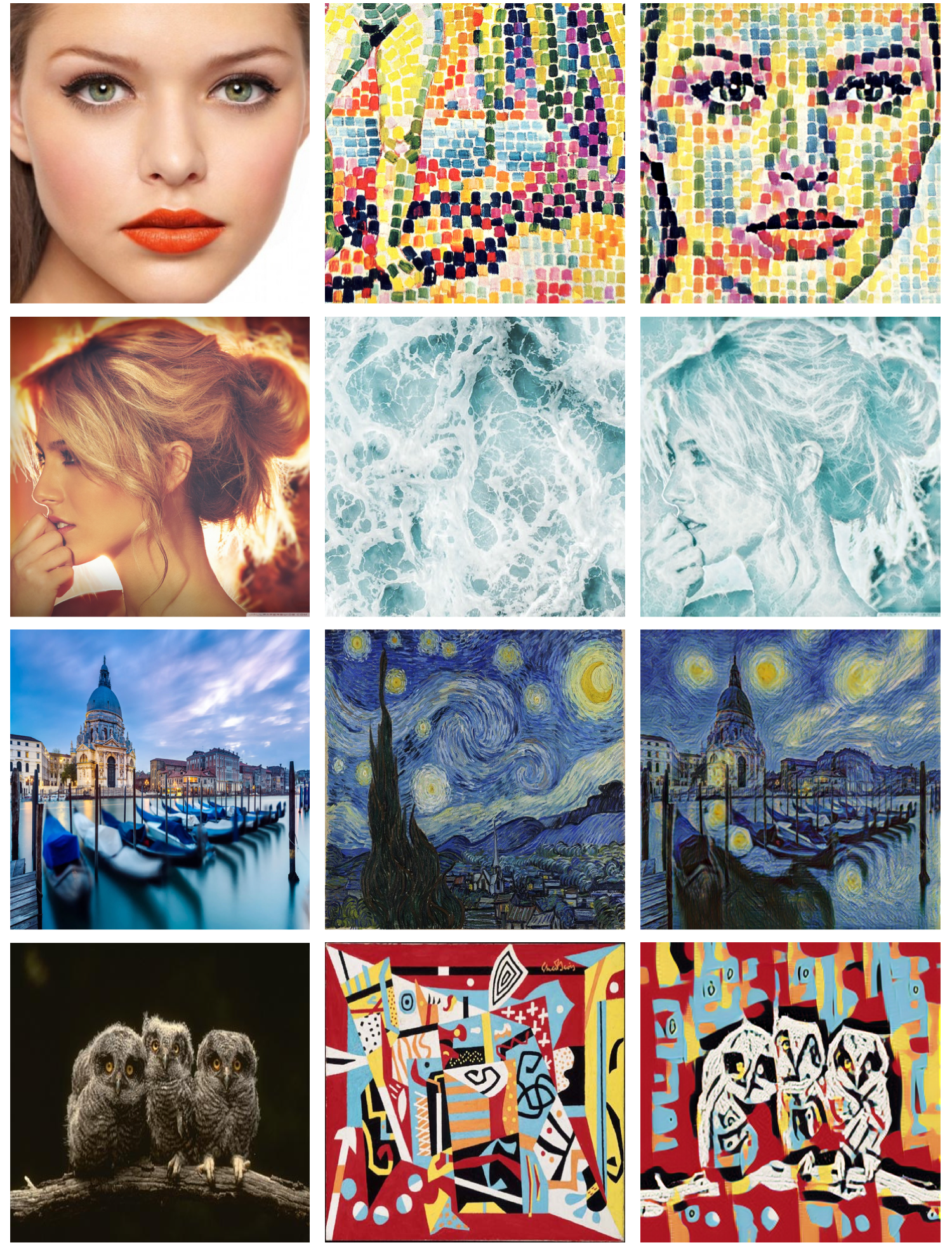}}
  \medskip
\end{minipage}
\end{center}
\caption{Qualitative results of high resolution stylization. Here, left images are content images, middle images are style images and right images are stylized images at 512 px.}
\label{fig:exp_large}
\end{figure*}

In supplementary material, we further demonstrate some stylization results on high resolution. Two Revision Networks are adopted at 256 px and 512 px successively to generate stylized images with 512 px.
As shown in Fig. \ref{fig:exp_large}, style patterns on multiple granularity are properly transferred by our method.

\end{document}